%% file: main.tex
\documentclass[runningheads, table, dvipsnames]{llncs}
\usepackage{graphicx}
\usepackage{tikz}
\usepackage{comment}
\usepackage{amsmath,amssymb} % define this before the line numbering.
\usepackage{color}

\usepackage[accsupp]{axessibility}  % Improves PDF readability for those with disabilities.

\input{macros}

\begin{document}
% \linenumbers
\pagestyle{headings}
\mainmatter
\def\ECCVSubNumber{5102}  % Insert your submission number here

\title{Weakly-supervised segmentation of\\ referring expressions}

\begin{comment}
\titlerunning{ECCV-22 submission ID \ECCVSubNumber} 
\authorrunning{ECCV-22 submission ID \ECCVSubNumber} 
\author{Anonymous ECCV submission}
\institute{Paper ID \ECCVSubNumber}
\end{comment}
%******************

\titlerunning{Weakly-supervised segmentation of\\ referring expressions}
\author{Robin Strudel \and
Ivan Laptev \and
Cordelia Schmid
}
\authorrunning{R. Strudel}
\institute{Inria, Ecole normale supérieure, CNRS,\\ PSL Research University, 75005
  Paris, France.
}

\maketitle

\begin{abstract}
Visual grounding localizes regions (boxes or segments) in the image
corresponding to given referring expressions.  In this work we address image
segmentation from referring expressions, a problem that has so far only been
addressed in a fully-supervised setting.  A fully-supervised setup, however,
requires pixel-wise supervision and is hard to scale given the expense of manual
annotation.  We therefore introduce a new task of weakly-supervised image
segmentation from referring expressions and propose Text grounded semantic
SEGgmentation (TSEG) that learns segmentation masks directly from image-level
referring expressions without pixel-level annotations.  Our transformer-based
method computes patch-text similarities and guides the classification objective
during training with a new multi-label patch assignment mechanism. The resulting
visual grounding model segments image regions corresponding to given natural
language expressions.  Our approach TSEG demonstrates promising results for
weakly-supervised referring expression segmentation on the challenging
\mbox{PhraseCut} and \mbox{RefCOCO} datasets.  TSEG also shows competitive
performance when evaluated in a zero-shot setting for semantic segmentation on
Pascal VOC.

\end{abstract}

\input{figures/teaser}

\section{Introduction}
\label{sec:intro}

Image segmentation is a key component for a wide range of applications including
virtual presence, virtual try on, movie post production and autonomous driving.
Powered by modern neural networks and supervised learning, image segmentation
has been significantly advanced by recent
work~\cite{chen18,cheng21,liu21_swin,strudel21}.  While most of this work
addresses semantic segmentation, the more general problem of visual grounding
beyond segmentation of pre-defined object classes remains open.  Moreover, the
majority of existing method assume full supervision and require costly
pixel-wise manual labeling of training images which prevents scalability.

Manual supervision has been recognized as a bottleneck in many vision tasks
including object detection~\cite{bilen2014weakly,kantorov2016contextlocnet,li2016weakly} and
segmentation~\cite{ahn18,araslanov20,fan18,zhou16}, text-image and text-video
matching~\cite{miech2020end,radford21} and human action
recognition~\cite{bojanowski2014weakly,ghadiyaram2019large}.  To this end,
self-supervised methods explore regularities in images and videos and learn
transferable visual representations without manual
supervision~\cite{chen20,doersch2015unsupervised}. Other weakly-supervised
methods exploit partial and possibly noisy supervision that is either
readily-available or less costly to
annotate~\cite{bilen2014weakly,miech2020end}. In particular, weakly-supervised
methods for image segmentation avoid the costly pixel-wise annotation and limit
supervision to image-level labels~\cite{ahn18,araslanov20,fan18,zhou16}. Such
methods, however, remain restricted to predefined sets of classes.

A referring expression is a short text describing a visual entity such
as \textit{man sitting on grass} or \textit{wooden stairway}, see Fig.~\ref{fig:teaser}. The task of referring expression segmentation~\cite{hu16,yu18}
generalizes image segmentation from pre-defined object classes to free-form
text. Given an input image and text queries (referring expressions), one should
generate image segments for each referring expression.  This enables
segmentation using compositional referring expressions such as \textit{man
sitting on grass} and \textit{wooden stairway}.
Despite the promise of scalability, existing approaches to referring expression
segmentation require pixel-wise annotation and, hence, remain limited by the
size of existing datasets.

Our work aims to advance image segmentation beyond limitations imposed by the
pre-defined sets of object classes and the costly pixel-wise manual annotations.
Towards this goal, we propose and address the new task of {\em weakly-supervised
referring expression segmentation}.   As this task comprises difficulties of the
weakly-supervised segmentation and referring expression segmentation, it
introduces new challenges.  In particular, existing weakly-supervised methods
for image segmentation typically rely on the completeness of image-level labels,
i.e., the absence of a car in the annotation implies its absence in the image.
This  completeness assumption does not hold for referring expression
segmentation. Furthermore, the vocabulary is open and compositional.

To address the above challenges and to learn segmentation from text-based
image-level supervision, we introduce a new global weighted pooling mechanism
denoted as Multi-label Patch Assignment (MPA).  Our method for Text grounded
semantic SEGgmentation (TSEG) incorporates MPA and extends the recent
transformer-based Segmenter architecture~\cite{strudel21} to referring
expression segmentation.  We validate our method and demonstrate its encouraging
results for the task of weakly-supervised referring expression segmentation on
the challenging PhraseCut~\cite{wu20} and RefCOCO~\cite{yu16} datasets. We also evaluate TSEG 
in a zero-shot setting for semantic segmentation and obtain competitive performance on the Pascal VOC dataset~\cite{everingham10}.  

In summary, our work makes the following three contributions.  (i)~We introduce
the new task of weakly-supervised referring expression segmentation and propose
an evaluation based on the PhraseCut and RefCOCO datasets.  (ii)~We propose
TSEG, a new method addressing weakly-supervised referring expression
segmentation with a multi-label patch assignment score.  (iii)~We demonstrate
advantages of TSEG through a number of ablations and experimental comparisons on
the challenging  PhraseCut and RefCOCO datasets. Furthermore, we demonstrate
competitive results for zero shot semantic segmentation on PASCAL VOC.

\section{Related Work}

\noindent\textbf{Weakly-supervised semantic segmentation.}  Given an image as
input, the goal of semantic segmentation is to identify and localize classes
present in the image, e.g. annotate each pixel of the input image with a class
label. Weakly-supervised Semantic Segmentation (WSS) has been introduced by
\cite{zhou16} and trains models using only image labels as supervision.
Zhou~\textit{et al.} \cite{zhou16} use Class Activation Maps (CAMs) of a Fully
Convolutional Network (FCN) combined with Global Average Pooling (GAP) to obtain
segmentation maps with a pooling mechanism.  As CAMs tend to focus on most
discriminative object parts~\cite{wei17}, recent methods deploy more elaborate
multi-stage approaches using pixel affinity~\cite{ahn19,ahn18}, saliency
estimation~\cite{fan19,fan18,huang18_dsrg,lee19_iccv,wang17,yu19} or seed and
expand strategies~\cite{huang18_dsrg,kolesnikov16,wei17}.

While these methods provide improved segmentation, they require multiple
standalone and often expensive networks such as saliency
detectors~\cite{fan18,huang18_dsrg,yu19} or segmentation networks based on
pixel-level affinity~\cite{ahn19,ahn18}.  Single-stage methods have been
developed based on multiple instance learning (MIL) \cite{pinheiro15} or
expectation-maximization (EM) \cite{papandreou15} approaches where masks are
inferred from intermediate predictions. Single-stage methods have been
overlooked given their inferior accuracy until the work of Araslanov~\textit{et
al.}~\cite{araslanov20} that proposed an efficient single-stage method
addressing the limitations of CAMs. Araslanov~\textit{et al.}~\cite{araslanov20}
introduces a global weighted pooling (GWP) mechanism which we extend in this
work with a new multi-label patch assignment mechanism (MPA). In contrast to
prior work on weakly-supervised semantic segmentation, TSEG is a single-stage
method that scales to the challenging task of referring expression segmentation.\\

\noindent\textbf{Referring expression segmentation.}
Given an image and a referring expression, the goal of referring expression
segmentation is to annotate the input image with a binary mask localizing the
referring expression.  A fully-supervised method~\cite{hu16} proposed to first
combine features of a CNN with a LSTM and then decode them with
a FCN.  To improve segmentation masks, \cite{yu18} uses a two-stage method based
on Mask-RCNN \cite{he17} features combined with a LSTM. To overcome the
limitation of FCN  to model global context and learn richer cross-modal
features, \textit{state-of-the-art} approaches \cite{ding21,hu20,ye19} use a
decoding scheme based on cross-modal attention. Despite their effectiveness,
these methods are fully-supervised which limits their scalability.  Several
weakly-supervised approaches tackle detection tasks such as referring expression
comprehension~\cite{chen18_knowledge,gupta20,liu19,liu21,xiao17} by enforcing
visual consistency \cite{chen18_knowledge}, learning language reconstruction
\cite{liu19} or with a contrastive-learning objective \cite{gupta20}. These
methods rely on an off-the-shelf object detector, Faster-RCNN \cite{ren17}, to
generate region proposals and are thus limited by the object detector accuracy.
None of these weakly-supervised methods address the problem of referring
expression segmentation which is the focus of our work.  TSEG is a novel
approach that tackles weakly-supervised referring expression segmentation based
on the computation of patch-text similarities with a new multi-label patch assignment mechanism (MPA). \\

\noindent\textbf{Transformers for vision and language.} Transformers
\cite{vaswani17} are now state of the art in many natural language processing
(NLP) \cite{devlin19} and computer vision
\cite{arnab21,cheng21,vit20,liu21_swin,strudel21} tasks.  Such methods capture
long-range dependencies among tokens (patches or words) with an attention
mechanism and achieve impressive results in the context of vision-language
pre-training at scale with methods such as CLIP \cite{radford21}, VisualBERT
\cite{li19}, DALL-E \cite{ramesh21} or ALIGN \cite{jia21}. Specific to referring
expressions, MDETR \cite{kamath21} recently proposed a method for visual
grounding based on a cross-modal transformer decoder trained on a
fully-supervised visual grounding task.  Several methods perform zero-shot
semantic segmentation with pre-trained fully supervision
models~\cite{ghiasi21,xu21,zabari21,zhou21_denseclip}.  Most similar to our
work, GroupViT \cite{xu22} relies on a large dataset of 30M image-text pairs to
learn segmentation masks from text supervision, but the objective function and
model architecture are different.

Our TSEG approach aims to learn patch-text associations while using only image-level
annotations with referring expression. TSEG
builds on CLIP~\cite{radford21} and uses separate encoders for different
modalities with a cross-modal late-interaction mechanism. Its segmentation
module builds on Segmenter~\cite{strudel21} which shows that interpolating patch
features output by a Vision Transformer (ViT)~\cite{vit20} is a simple and
effective way to perform semantic segmentation. Here, we extend this work to
perform cross-modal segmentation.  TSEG leverages a novel patch-text interaction
mechanism to compute both image-text matching scores and pixel-level
text-grounded segmentation maps in a single forward pass.

\section{Method}
\input{figures/overview}

TSEG takes as input an image and a number of referring expressions and outputs
a confidence score (Fig.~\ref{fig:overview}, top-right) along with a
segmentation mask (Fig.~\ref{fig:overview}, bottom-right) for each referring
expression. During training no segmentation masks are available and image-level
labels are used to train  referring expression segmentation
(Fig.~\ref{fig:overview}, top-right). TSEG is based on image patch-text matching
(Fig.~\ref{fig:overview} left).  An image encoder maps the input image to a
sequence of patch tokens and a text encoder maps each input referring expression
to a single text token. The tokens are then projected to a common embedding
space and patch-text cosine similarities are computed as described in Section
\ref{sec:encoders}. To obtain an image-level score for each referring
expression, the patch-text similarity matrix is summarized along the patch
dimension.  To do so, we introduce a novel multi-label patch assignment (MPA)
mechanism described in Section \ref{sec:global_pooling}. The model is then
trained end-to-end to predict the corresponding image-text pairs as described in
Section \ref{sec:train_inf}. At inference, the patch-text matrix is simply
interpolated for patches  to obtain pixel-level masks as described in Section
\ref{sec:train_inf}.   The choice of an appropriate global pooling mechanism is
important to learn accurate segmentation maps as illustrated in Figure
\ref{fig:patch_matching}. We evaluate its impact in Section
\ref{sec:experiments} and show that the novel multi-label patch assignment
mechanism outperforms existing ones by a significant margin.

\subsection{Patch-text similarity matrix}
\label{sec:encoders}

In this section we describe how to compute the similarity matrix between patches
of an image and several referring expressions. We consider an image represented
by $N$ patches $p_1, ... , p_N$ and a set of $L$ referring expressions
$t_1, ..., t_L$. Patches are encoded by tokens $(\mbf{x}_1, ..., \mbf{x}_N)$,
each referring expression consists of several words and is encoded by one token
$(\mbf{y}_1, ..., \mbf{y}_L)$. The resulting similarity matrix is
$\mbf{S} = (\mbf{x}_{i}\cdot \mbf{y}_{j})_{i,j} \in \mathbb{R}^{N \times L}$.
See Figure~\ref{fig:overview} left.\\

\noindent\textbf{Image encoder.} An image $I\in \mathbb{R}^{H \times W \times C}$ is split into a
sequence of patches of size $(P,P)$. Each image patch is then linearly
projected and a position embedding is added to produce a sequence of patch
tokens
$(p_1, ..., p_N) \in \mathbb{R}^{N \times D_{I}}$
where $N = HW/P^2$ is the number of patches, $D_{I}$ is the number of features.
A transformer encoder maps the input
sequence to a sequence of contextualized patch tokens
$(\mbf{x}_{1},..., \mbf{x}_{N}) \in \mathbb{R}^{N \times D_{I}}$.
See more details in Section~\ref{sec:implem}.

\noindent\textbf{Text encoder.} For each referring expression $t_{j}$, which can
consist of multiple words, we extract one token $\mbf{y}_j$. To do so the text
$t_j$ is tokenized into words using lower-case byte pair encoding (BPE)
\cite{bpe16} and [BOS], [EOS] tokens are added to the beginning and the end of
the sequence. A sequence of position embedding is added and a transformer
encoder maps the input sequence to a sequence of contextualized word tokens from
which the [BOS] token is extracted to serve as a global text representation
$\mbf{y}_j \in \mathbb{R}^{D_{T}}$.

\noindent\textbf{Patch-text similarity scores.} The visual and textual tokens
are linearly projected to a multi-modal common embedding space and
$L^2$-normalized. From the patch tokens $(\mbf{x}_{1},...,\mbf{x}_{N})$ and the global text tokens
$(\mbf{y}_{1},...,\mbf{y}_{L})$, we compute patch-text cosine similarities as
the scalar product and obtain the similarity matrix
\begin{equation}
  \mbf{S} = (s_{i,j})_{i,j} = (\mbf{x}_i \cdot \mbf{y}_{j})_{i,j},
\end{equation}
with  $\mbf{S} \in \mathbb{R}^{N \times L}$.
The similarities are in the range $[-1, 1]$ and scaled with a
learnable temperature parameter $\tau > 0$ controlling their range.

\subsection{Global Pooling Mechanisms}
\label{sec:global_pooling}
\input{figures/patch_matching}

To leverage image-level text supervision, we need to map the matrix
$\mbf{S}\in \mathbb{R}^{N\times L}$ of patch-text similarities to an image-level
score for each referring expression, i.e.,  $\mbf{z} \in \mathbb{R}^{L}$.  The
score vector $\mbf{z}$ allows us to compute a classification loss using ground
truth referring expressions.  Note that we cannot compute per-pixel losses given
the lack of pixel-wise supervision in weakly-supervised settings.\\

\textbf{Global average and max pooling (GAP-GMP).}
A straightforward way of pooling is global average pooling (GAP), where we
average the similarities for a given referring expression over all patches of an
image:

\begin{equation}
  z^{GAP}_{j} = \frac{1}{N}\sum_{i=1}^{N}s_{i, j}.
\end{equation}
This score is expected to be high if the referring expression is contained in
the image. However, the score is dependent on the object size and results in low
scores  for small objects.  An alternative to GAP is global max pooling (GMP):
\begin{equation} z^{GMP}_{j} = \max_{i}(s_{i, j}).
\end{equation} The max operation in GMP decreases the influence of the object
size, however it tends to focus on most discriminative regions of a class
\cite{wei17}.\\

\textbf{Global weighted pooling (GWP).}
To address the shortcomings of GAP and GMP, we follow~\cite{araslanov20} and
make use of weighted pooling.  Global weighted pooling replaces the constant
patch weights $1/N$ in the sum of $GAP$ by weights
$\mbf{W} = (w_{i,j})_{i, j} \in \mathbb{R}^{N \times L}$.  The final score of a
referring expression is then the weighted average of similarities:
\begin{equation}
  \label{eq:gwp}
  z_{j}^{GWP}= \sum_{i=1}^{N}w_{i, j}s_{i, j},
\end{equation}
as illustrated in Figure \ref{fig:patch_matching} left.
In practice, $\mbf{W}$ is  defined  in terms of spatially normalized mask scores
%is defined as the normalization of masks
$\mbf{M} = (m_{i,j})_{i,j} \in \mathbb{R}^{N \times L}$,
based on $w_{i,j}=m_{i, j}/(\sum_{i}m_{i,j}+\varepsilon)$
where $\varepsilon > 0$ allows for $\sum_i w_{i,j} = 0$ when mask scores are below
a threshold. GAP is a particular case of GWP where $m_{i,j}=1$ for all $i,j$ and
$\varepsilon=0$.
We next describe two methods to compute masks $\mbf{M}$ from the similarity matrix  $\mbf{S}$.\\

\noindent\textbf{Masks by single-label patch assignment
(SPA)~\cite{araslanov20}.} We aim at assigning patches to the relevant referring
expression. To do so, we apply a softmax operation over all referring
expressions $(\mbf{y}_1,...,\mbf{y}_L)$ for each patch $\mbf{x}_{i}$:
\begin{equation}
  \label{eq:spa}
  m_{i, j}^{SPA}= \frac{e^{s_{i,j}}}{e^{s_{bg}} +\sum_{j=1}^{L}e^{s_{i,j}}}.
\end{equation}
We add a background column $(s_{i, 0})_i$ and assign it a constant equal to
$s_{bg} = 0$ for all patches $\mbf{x}_{i}$. This allows to assign patches with
low scores $s_{i, j} < 0$ to the background.  The masks are then soft
assignments with $\sum_{j=0}^{L} m_{i,j} = 1$ for any patch $i$. This patch
assignment can be viewed as multi-class classification which is typical for
semantic segmentation where one pixel is matched to a \textit{single label} as proposed
by \cite{araslanov20}.

This single-label patch assignment (SPA) is illustrated in Figure
\ref{fig:patch_matching} center.  The softmax operation over referring
expressions softly enforces the correspondence  of a patch to one expression.
However, this definition is problematic for referring expression segmentation
where the masks of several expressions can overlap. We illustrate this in Figure
\ref{fig:patch_matching} center where pixels corresponding to \textit{jumping man}
and \textit{legs} have lower mask weights on the overlapping region. Such lower mask
weights result in decreased image-level scores for both of the expressions.\\

\noindent\textbf{Masks by multi-label patch assignment (MPA).}  We propose
multi-label patch assignment (MPA) that overcomes the above limitations of SPA.
For each patch $\mbf{x}_{i}$, we rely on binary classification between a
referring expression $\mbf{y}_{j}$ and the background based on:

\begin{equation}
  \label{eq:mpa}
  m_{i,j}^{MPA}= \frac{e^{s_{i,j}}}{e^{s_{bg}}+e^{s_{i,j}}}.
\end{equation}
In this case, each patch can be assigned to multiple referring expressions, see
Figure \ref{fig:patch_matching} right. The masks are not mutually exclusive and
each referring expression can be assigned a score $m_{i,j}^{MPA} \in [0, 1]$
without softmax imposed constraints. Patch assignment is viewed as a multi-label
classification problem, this property is highly beneficial when performing
weakly-supervised referring expression segmentation, as shown in Section
\ref{sec:experiments}.\\

\noindent\textbf{Image-text scores.}
We compute GWP scores $\mbf{z}^{GWP}$ with (\ref{eq:gwp}) using the masks
$\mbf{M}$ defined according to one of the assignment mechanism defined in
(\ref{eq:spa}),(\ref{eq:mpa}). Then, we compute mask size scores
$\mbf{z}^{size}$ as
\begin{equation}
  \label{eq:size_gwp}
  z^{size}_{j} = (1-\overbar{m}_{j})^{p}\log(\lambda+\overbar{m_{j}}),
\end{equation}
with $\overbar{m_{j}}=\frac{1}{N}\sum_{i=1}^{N}m_{i,j}$. This $\mbf{z}^{size}$
is a size-penalty term introduced by \cite{araslanov20} to enforce mask
completeness, e.g. $z_{j}^{size} < 0$ for small masks.  The magnitude of this
penalty is controlled by $\lambda$. Due to the normalization, $\mbf{W}$ used in
GWP is invariant to the masks size $\mbf{M}$ and $\mbf{z}^{size}$ enforces masks
to be complete.  The final score defining the presence of a referring expression
$t_j$ in the image is defined as the sum:
\begin{equation}
z_{j} = z^{GWP}_{j}+z^{size}_{j}.
\end{equation}

\subsection{Training and inference}
\label{sec:train_inf}

In the following we describe our weakly supervised and fully supervised training
procedure. Furthermore, we present the approach used for inference.

\noindent\textbf{Weakly-supervised learning.} Weakly-supervised segmentation is
usually addressed on datasets with a fixed number of classes. To handle the more
general case where visual entities in the image are defined by referring
expressions we use referring expressions of samples in a mini-batch as positive
and negative examples.  Given a mini-batch containing (image, referring
expression) pairs, the model has to predict the subset of referring expressions
present in each image.  For each image, we extract image-text scores
$\mbf{z} \in \mathbb{R}^{L}$ from the similarity matrix $\mbf{S}$ using one of
the pooling mechanism described in the previous section.  Finally, we optimize
over the scores to match  ground truth pairings $\overbar{\mbf{z}}$ with the
multi-label soft-margin loss function \cite{ahn18,araslanov20,wei18} as a
classification loss,
\begin{eqnarray*}
  \mathcal{L}_{cls}(\mbf{z}, \overbar{\mbf{z}}) =
  \sum_{j=1}^{L}
  -\overbar{z}_{j} \log\left(\sigma(z_{j})\right)
  -(1-\overbar{z}_{j})\log\left(\sigma(-z_{j})\right),
\end{eqnarray*}
where $\sigma(x) = 1/(1+\exp(-x))$ is the sigmoid function. The loss encourages $z_{j} > 0$ for positive image-text pairs and
$z_{j} < 0$ for negative pairs.
\smallskip

\noindent\textbf{Fully-supervised learning.}  In the fully-supervised case,
segmentation is learned from a dataset of images annotated with referring
expressions and their corresponding segmentation masks. Only positive referring
expressions $(\textbf{y}_1,...,\textbf{y}_L)$ are passed to the text encoder and the similarity
matrix $\mbf{S}$ is bilinearly interpolated to obtain pixel-level similarities
of shape $\mathbb{R}^{H\times W \times L}$. Then, we minimize the Dice loss
between the sigmoid of the pixel-level similarities $\mbf{M} = \sigma(\mbf{S})$
and the ground truth masks $\overbar{\mbf{M}}$:
\begin{equation}
  \mathcal{L}_{dice}(\mbf{M}, \overbar{\mbf{M}}) = 1-2\frac{|\mbf{M}\cap\overbar{\mbf{M}}|}{|\mbf{M}|+|\overbar{\mbf{M}}|},
\end{equation}
where $|\mbf{M}| = \sum_{i, j} m_{i,j}$ and $\mbf{M}\cap\overbar{\mbf{M}} = (m_{i,j}\overbar{m}_{i,j})_{i,j}$.\smallskip

\noindent\textbf{Inference.} To produce segmentation masks, we reshape the
patch-text masks $\mbf{M} \in \mathbb{R}^{N \times L}$ into a 2D map and
bilinearly interpolate it to the original image size to obtain pixel-level masks
of shape $\mathbb{R}^{H\times W \times L}$. For SPA, pixel annotations are
obtained by adding a background mask to $\mbf{M}$ and applying an argmax over
the refering expressions. For MPA, we threshold the values of $\mbf{M}$ using
the background score.  For GAP and GMP, we follow the standard approach
from~\cite{ahn18} to compute the masks $\mbf{M}$. Directly interpolating
patch-level similarity scores to generate segmentation maps has been proven
effective by Segmenter \cite{strudel21} in the context of semantic segmentation.
Our decoding scheme is an extension of Segmenter linear decoding where the set
of fixed class embeddings is replaced by text embeddings.

\section{Experiments}
\label{sec:experiments}

In this section we first outline datasets and implementation details in
Sections~\ref{sec:dataset} and \ref{sec:implem}.  We then  validate our
implementation of two \textit{state-of the-art} methods for weakly-supervised
semantic segmentation in Section~\ref{sec:wseg_pascal} .  Next, we ablate
different parameters of the proposed TSEG method for the task of referring
expression segmentation in Section~\ref{sec:tseg_ablations}.  Finally, we
compare TSEG to methods introduced in Section~\ref{sec:wseg_pascal} on referring
expression datasets in Section~\ref{sec:wseg_re}.

\subsection{Datasets and metrics}
\label{sec:dataset}

\noindent\textbf{Pascal VOC 2012.} Pascal~\cite{everingham10} is an
established benchmark for weakly-supervised semantic segmentation. Following
standard practice \cite{ahn19,ahn18,araslanov20,kolesnikov16}, we augment the
original training data with additional images from \cite{hariharan11}. The
dataset contains 10.5K images for training and 1.5K images for validation.\\

\noindent\textbf{PhraseCut.} PhraseCut~\cite{wu20} is the largest  referring
expression segmentation dataset with 77K images annotated with 345K referring
expressions from Visual Genome \cite{krishna17}. The expressions comprise a wide
vocabulary of objects, attributes and relations.
The
dataset is split into 72K images, 310K expressions for training and 3K images,
14K expressions for validation.\\

\noindent\textbf{RefCOCO.} RefCOCO and RefCOCO+~\cite{yu16} are the two most
commonly used datasets for referring expression segmentation and comprehension.
RefCOCO has 20K images and 142K referring expressions for 50K objects while
RefCOCO+ contains 20k images and 142K expressions for 50K objects.  RefCOCO+ is
a harder dataset where words related to the absolute location of the objects are
forbidden. RefCOCOg is a dataset of 27K images with 105K expressions referring
to 55K objects. Compared to RefCOCO(+), RefCOCOg has longer sentences and richer
vocabulary.\\

\noindent\textbf{Metrics.} We follow previous work and report mean Intersection
over Union (mIoU) for all Pascal classes. For referring expression segmentation
we use standard metrics where mIoU is the IoU averaged over all image-region
pairs resulting in a balanced evaluation for small and large
objects~\cite{yu16,wu20}.

\subsection{Implementation details}
\label{sec:implem}

\noindent\textbf{Initialization.} Our TSEG model contains an image encoder
initialized with an ImageNet pre-trained Vision
Transformer~\cite{vit20,steiner21} and a text encoder initialized with a
pre-trained BERT model~\cite{devlin19}.  We use ViT-S/16 \cite{steiner21} and
BERT-Small \cite{turc19} which are both expressive models achieving strong
performance on vision and language tasks, while remaining fast and compact. Our
model has a total number of 42M parameters. Following \cite{vit20,strudel21}, we
bilinearly interpolate ViT position embeddings when using an image resolution
that differs from its pre-training.\\

\noindent\textbf{Optimization.}
For weakly-supervised learning, we use SGD optimizer~\cite{robbins1951} with a
base learning rate $\gamma_{0}$, and set weight decay to $10^{-4}$.  Following
DeepLab~\cite{chen18}, we adopt the poly learning rate decay
$\gamma = \gamma_{0}(1-\frac{n_{iter}}{n_{total}})^{0.9}$.  We use a stochastic
drop path rate \cite{huang16} of 0.1 following standard practices to train
transformers \cite{devlin19,vit20,steiner21}.  For Pascal, PhraseCut and
RefCOCO, we set the base learning rate $\gamma_{0}=10^{-3}$.  We found this
learning scheme to be stable resulting in  good results for all three datasets.
Regarding training iterations and the batch size, we use 16K iterations and
batches of size 16 for Pascal, 80K iterations and batches of size 32 for
RefCOCO, and 120K iterations with batches of size 32 for PhraseCut.
When training on referring expressions, we randomly sample three positive
expressions per image on average.  The resolution of images at train time is set
to $384 \times 384$ and following standard practices we use random rescaling,
horizontal flipping and random cropping.

For the fully-supervised setup we use AdamW~\cite{kingma14,loshchilov19}
optimizer and set the base learning rate $\gamma_{0}$ to $5 \times 10^{-5}$. We
set the batch size to 16 for all datasets and use the same number of iterations
as for weakly-supervised setups. The resolution of images at train time is
$512 \times 512$.

\subsection{State-of-the-art methods for weakly-supervised semantic segmentation}
\label{sec:wseg_pascal}

\input{tables/wseg_pascal}

As we are the first to propose an approach for weakly-supervised learning for
referring expression segmentation, we implemented state-of-the-art methods for
weakly-supervised semantic segmentation to use as baselines. We use three
single-stage methods presented in Section~\ref{sec:global_pooling}, namely GMP
\cite{zhou16}, the seminal work GAP~\cite{ahn18}, and the more recent
state-of-the-art approach SPA~\cite{araslanov20}. SPA performs close to the best
two-stage  weakly-supervised methods, DRS \cite{kim21} and EPS \cite{lee21}, two
more complex  methods relying on off-the-shelf saliency detectors, which is not
the focus of our work.\\

Table \ref{tab:wseg_pascal} reports the performance on the Pascal VOC 2012 dataset. With a language model as 
class encoding as shown in Figure \ref{fig:overview}, we obtain similar
performances as GAP \cite{ahn18} and SPA \cite{araslanov20} using the same
WideResNet38 backbone.  By using the more recent ViT-S/16 backbone with SPA,
we obtain 66.4\% mIoU, a 4\% gain over WideResNet38.  We also report results
with GMP \cite{zhou16} for which we did not find methods reporting results on
Pascal VOC 2012. The GMP results are below the GAP results and again the
ViT-S/16 backbone gives improved results.  In the following sections we use
ViT-S/16 as the image encoder, BERT-Small as the text encoder and GAP, GMP and
SPA as a point of comparison to our proposed TSEG method. The models can
directly be used to perform referring expression segmentation by replacing the
class label given as input to the language model by referring expressions.

\subsection{TSEG ablations}
\label{sec:tseg_ablations}
\input{tables/tseg_ablations}

We now perform weakly-supervised referring expression segmentation. At train
time the model has to maximize the score of the image and text embeddings of
correct pairings while minimizing the score of incorrect pairings. At test time,
following the standard visual grounding setting, the model is given as input the
set of referring expressions present in the image and outputs a mask for each
referring expression. TSEG uses the proposed MPA to compute scores from
patch-text similarities.\\

Table \ref{tab:ablations} reports ablations of our TSEG model on the PhraseCut
validation set. First, we ablate over the size penalty parameters $\lambda$ and
$p$ from Eq. \ref{eq:size_gwp} in Table \ref{tab:ablations}a.  Smaller $\lambda$
values induce a larger penalty for masks with a small size and larger $p$ values
increase the focal penalty term, see \cite{lin20} for more details.  
We find TSEG is quite robust to the objective hyperparameters $\lambda$ and $p$.
The best values are $\lambda=0.01$ and $p=5$; we fix $\lambda$
and $p$ to these values in the remaining of the paper.
Table \ref{tab:ablations}b reports performance for different cross-modal
embedding dimension, increasing the embedding size improves results overall. In
Table \ref{tab:ablations}c, we consider different definitions for the ground
truth. In the \textit{identity} setup, two referring expressions of a batch are
considered the same if they exactly match.  In the \textit{tf-idf} setup, the
similarity between two referring expressions if computed according to a  tf-idf
score. If a \textit{tabby cat} is present in an image, and
there is a \textit{brown cat} in a second image, the ground truth score for
\textit{brown cat} in the first image will be positive because both referring
expression share the word \textit{cat}. Using tf-idf performed slightly worse
than the identity score and we thus use \textit{identity} to define the ground
truth.  Table \ref{tab:ablations}d reports the validation score for an
increasing training dataset size. We observe that TSEG improves with the dataset
size, a desirable property of weakly-supervised segmentation approach where
annotations are much cheaper to collect than in the fully-supervised case.
Finally, Table \ref{tab:ablations}e reports results when pretraining the visual
backbone on only ImageNet for classification or by additionally pretraining the
visual and language model on RefCOCO for visual grounding. For pretraining on
COCO we use box ground truth annotations as follows. The model is given as input
an image and referring expressions to detect, for each referring expressions the
model predicts patches that are within the object bounding box.  We observe that
leveraging detection related information as pretraining improves the result by
3\%. In the following we report results with ImageNet pretraining only,
following standard practice from the weakly-supervised semantic segmentation
literature.

\subsection{Weakly supervised referring expression segmentation}
\label{sec:wseg_re}
\input{figures/methods}
\input{tables/wseg_re}

We now compare TSEG on referring expression datasets to
weakly supervised \textit{state-of-the-art} methods presented in Section \ref{sec:wseg_pascal}, we
report results in Table \ref{tab:wseg_re} and show qualitative results in Figure
\ref{fig:methods}.\\

\noindent\textbf{PhraseCut:} GMP and GAP achieve an mIoU of 5.7 and 9.3
respectively, showing that it is possible to learn meaningful masks using
referring expressions as labels. However, GAP averages patch-text similarity
scores and depends on the instance mask size which tends to generate
over-saturated activation maps (Fig.~\ref{fig:methods}a). GMP exhibits
complementary properties focusing on the most discriminative object parts
(Fig.~\ref{fig:methods}b).  SPA outperforms GAP and GMP with a mIoU of 21.1,
consistent with results on Pascal VOC 2012, see Table \ref{tab:wseg_pascal}.
MPA further improves SPA by 7\%, with 28.77\% mIoU on Phrasecut, showing its
crucial importance for referring expression segmentation.  This improvement can
partly be explained by the fact that our objective allows multiple masks to
overlap by design, a highly desirable property that is not satisfied by GMP, GAP
and SPA.  From Figure~\ref{fig:methods}d we observe that MPA generates more
complete masks with both higher recall, e.g. the \textit{thumb on bun} instance
is detected, and we obtain higher precision, e.g. masks achieve better
completeness as for the \textit{sitting woman} instance. Using CRF
\cite{chen_deeplab} further improves the performance to 30.12 mIoU. Qualitative
results are presented in Figure \ref{fig:qualitative}.\\

\input{figures/qualitative}

To obtain an upper-bound, we also train TSEG with full supervision and obtain a
49.6 mIoU. This is close to the best fully supervised method MDETR
\cite{kamath21}, which obtains 53.1 mIoU while pretraining on a much large
dataset annotated for visual grounding and higher training resolution. While
there is still a gap compared to full supervision,  we believe our proposed
results to be promising and the first step towards large-scale weakly supervised
referring expression segmentation. Additional qualitative results and comparison
  to the fully-supervised model are presented in the appendix.\\

\noindent\textbf{RefCOCO:} We also evaluate our method on the three RefCOCO
datasets and report results on the \textit{val} split in Table \ref{tab:wseg_re}.
Again, MPA outperforms GMP, GAP and SPA by a large margin. Training TSEG with full
supervision we obtain 66.00 mIoU on RefCOCO, 55.35 on RefCOCO+ and 54.71 on
RefCOCOg. This is slightly better than the best fully supervised method VLT
\cite{ding21}, which obtains 65.65, 55.50 and 52.99 mIoU respectively. There is
a larger gain from using full supervision than on PhraseCut. This could be
explained by more fine-grained  referring expressions such as \textit{broccoli
stalk that is pointing up and is touching a sliced carrot} or \textit{a darker
brown teddy bear in a row of lighter teddy bears} that are harder to localize
without pixel-level supervision.

\input{figures/zero_shot_pascal}

\subsection{Zero-shot transfer on Pascal VOC}
\label{sec:wseg_zeroshot}

We evaluate the ability of TSEG to detect and localize visual concepts from text
supervision by performing zero-shot experiments on Pascal VOC 2012 dataset, see
Fig.~\ref{fig:zero_shot}. We take our TSEG model trained on the PhraseCut
dataset, i.e., with the text supervision based on the referring expressions from
PhraseCut.  We, then, pass the names of Pascal classes as input to the text
encoder and obtain segmentation masks and confidence scores for all 20 object
classes in each image. We filter classes by thresholding with the model
confidence scores then use argmax between the remaining masks to determine the
class of each pixel. We set the threshold to 0.5.\\

In the zero-shot setting, our TSEG model achieves an mIoU of 48.5 while the SPA
baseline achieves an mIoU of 43.5.  Interestingly, TSEG performs well on all
classes except the \textit{person} class. As can be observed from
Figure~\ref{fig:zero_shot_fail}, the model does not detect the \textit{person}
label, but can be improved with label engineering by using more specific labels
for the text encoder, such as \textit{woman} and \textit{rider}.  This bias
partly comes from the annotations of PhraseCut training set and we believe that
the need for label engineering may be reduced by training TSEG on a larger
dataset with richer text annotations. On the person class, by passing
\textit{person} as input to the text encoder we obtain an IoU of 0.6 while by
merging masks for the words \textit{man, woman, men, women, child, boy, girl,
baby} we improve the IoU to 30.4. By performing label engineering, TSEG reaches
50.3 mIoU. In comparison, GroupViT~\cite{xu22} reports an mIoU of
51.2, but  it has been trained on a much larger dataset of 30M image-text pairs
and was designed for zero-shot segmentation. TSEG performs comparably to
GroupViT, while trained on 350k image-text pairs.  This demonstrates the ability
of our approach to learn general visual concepts accurately.

\input{figures/zero_shot_fail}

\section{Acknowledgements}
This work was partially supported by the HPC resources from GENCI-IDRIS  (Grant
2021-AD011011163R1), the Louis Vuitton ENS Chair on Artificial Intelligence, and
the French government under management of Agence Nationale de la Recherche as
part of the ”Investissements d’avenir” program, reference ANR-19-P3IA-0001
(PRAIRIE 3IA Institute).

\section{Conclusion}

This work introduces TSEG for weakly-supervised referring expression
segmentation.  We propose a multi-label patch assignment (MPA) mechanism that
improves previous methods by a margin on this task.  We believe our work makes
an important step towards scalable image segmentation from natural language.
Future work will address how to reduce the performance gap between weakly
supervised and fully supervised methods and segment regions directly from
image captions.

\clearpage

\bibliographystyle{splncs04}
\bibliography{egbib}

\clearpage
\section{Appendix}

\textbf{Qualitative results.}
We present additional qualitative results in
Figures~\ref{fig:supmat_weak_vs_full} and \ref{fig:supmat_weak}.
In particular, we compare TSEG trained with weak supervision to the same
model trained with full supervision in Figure~\ref{fig:supmat_weak_vs_full}.
TSEG captures cloth related concepts, animals and parts of the bodies
reasonably well, however it can fail at capturing colors, distinguish between
a book and a laptop, or between a blue jean and different type of trousers.
In Figure~\ref{fig:supmat_weak}, we observe that TSEG captures a
rich variety of visual concepts, even rarely occurring ones quite accurately.

\vspace{-0.2cm}
\input{supmat/weak_vs_full}
\input{supmat/weak}
\end{document}

%% file: macros.tex
\usepackage{amsmath}
\usepackage{amssymb}
\usepackage{amsfonts}

\newcommand{\overbar}[1]{\mkern 1.5mu\overline{\mkern-1.5mu#1\mkern-1.5mu}\mkern 1.5mu}

\usepackage{algorithm}% http://ctan.org/pkg/algorithm
\usepackage{algpseudocode}% http://ctan.org/pkg/algorithmicx

\usepackage{epsfig}
\usepackage{graphicx}
\usepackage{subcaption}

\usepackage{booktabs}
\usepackage{pifont}
\newcommand{\cmark}{\ding{51}}%
\newcommand{\xmark}{\ding{55}}%

\usepackage{multicol}
\usepackage{multirow}
\newcommand{\tsdagger}{{\textsuperscript{\textdagger}}}

\newcommand{\mbf}[1]{\mathbf{#1}}

\definecolor{LightCyan}{rgb}{0.88,1,1}

\newcommand{\tablestyle}[2]{\setlength{\tabcolsep}{#1}\renewcommand{\arraystretch}{#2}\centering\footnotesize}

\newlength\savewidth\newcommand\shline{\noalign{\global\savewidth\arrayrulewidth
  \global\arrayrulewidth 1pt}\hline\noalign{\global\arrayrulewidth\savewidth}}

%% file: figures/teaser.tex
\begin{figure}[t]
  \centering
  \includegraphics[width=1.0\textwidth]{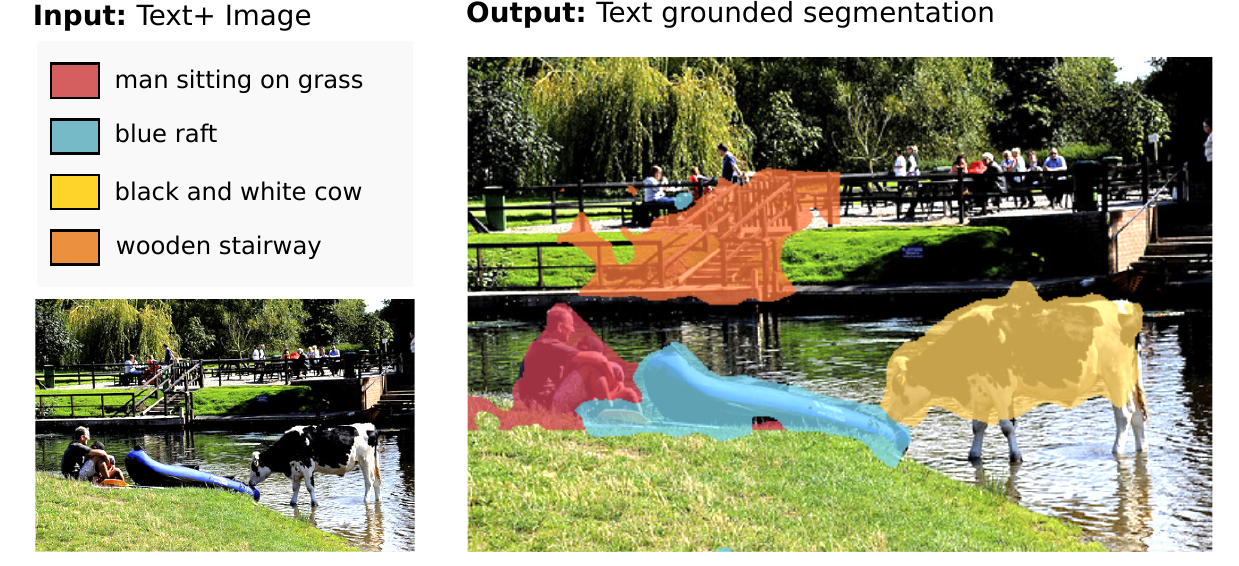}
  \vspace{-.3cm}
  \caption{Given an image and a set of referring expressions such as \textit{man sitting
      on grass} and \textit{wooden stairway}, TSEG segments the image regions corresponding to the input expressions. Here we show results of our approach TSEG for a test image of the PhraseCut dataset.
      Contrary to other existing methods, TSEG only uses image-level referring expressions during training and hence does not require pixel-wise supervision.
  \vspace{-.4cm}
  }
  \label{fig:teaser}
\end{figure}

%% file: figures/overview.tex
\begin{figure*}[t]
  \centering
  \vspace{-0.4cm}
  \includegraphics[width=\textwidth]{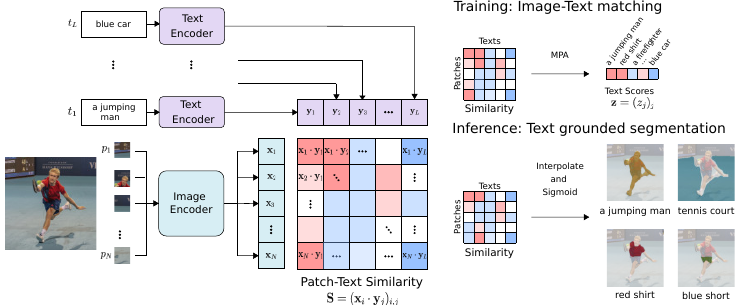}
   \vspace{-.4cm}
  \caption{Overview of our approach TSEG. (Left) Image patches and referring
    expressions are mapped with transformers to patch and text embeddings and then compared
    by computing patch-text cosine similarity scores.
    (Right - Training) Our global pooling mechanism with multi-label patch assignment (MPA) reduces
    patch-text similarity scores to image-level labels to train the model for referring expression classification.  
    (Right - Inference) Sequences of patch scores (columns) are rearranged into 2D masks and bilinearly interpolated to obtain pixel-level referring expression masks.
  }
  \label{fig:overview}
  \vspace{-0.4cm}
\end{figure*}

%% file: figures/patch_matching.tex
\begin{figure*}[t]
  \centering
  \includegraphics[width=\textwidth]{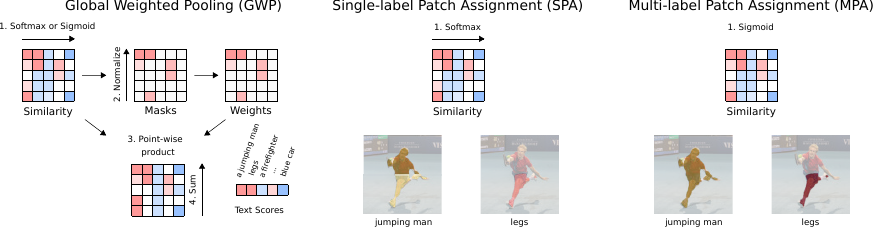}
  \vspace{-.5cm}
  \caption{(Left)
    A patch assignment mechanism computes masks from patch-text similarities, the
    masks are used as weights in the global weighed pooling. (Center) SPA:
    assignment with a softmax on text channels, softly enforcing a single label
    per patch. (Right) MPA: assignment with a sigmoid, generalizing to multiple labels per patch.
    }
    \vspace{-.2cm}
  \label{fig:patch_matching}
\end{figure*}

%% file: tables/wseg_pascal.tex
\begin{table}[t]
\small
\setlength\tabcolsep{3pt}
\centering
\begin{tabular}{llccc}
  \toprule
  Method & Image encoder & \multicolumn{2}{c}{Class encoding} & mIoU \\
   & & Vector & Language model & \\
  \midrule
  % DeepLabv3+ \cite{chen18} & WideResNet38 & \cmark & \xmark & \xmark & 80.8 \\
  % GMP [?] & ? & \xmark & \xmark & \cmark & ? \\
  GAP \cite{ahn18} & WideResNet38 & \cmark & \xmark & 48.0 \\
  GAP \cite{ahn18}\tsdagger & WideResNet38 & \xmark & \cmark & 46.8 \\
  GAP \cite{ahn18}\tsdagger & ViT-S/16 & \xmark & \cmark & 50.2 \\
  \hline
  GMP \cite{zhou16}\tsdagger & WideResNet38 & \xmark & \cmark & 44.3 \\
  GMP \cite{zhou16}\tsdagger & ViT-S/16 & \xmark & \cmark & 48.1 \\
  \hline
  SPA \cite{araslanov20} & WideResNet38 & \cmark & \xmark & 62.7 \\
  SPA \cite{araslanov20}\tsdagger & WideResNet38 & \xmark & \cmark & 62.4 \\
  SPA \cite{araslanov20}\tsdagger & ViT-S/16 & \xmark & \cmark & \textbf{66.4} \\
\bottomrule
\end{tabular}
\caption{State-of-the-art single-stage methods for weakly-supervised semantic
  segmentation on the Pascal VOC validation set. \tsdagger \, denotes our implementation. Multi-scale processing and CRF are used for inference.}
\label{tab:wseg_pascal}
% \vspace{-0.4cm}
\end{table}

%% file: tables/tseg_ablations.tex
\begin{table*}[t]
% \vspace{-1.5cm}
\centering
\subfloat[Size penalty term.]{
\begin{minipage}{0.5\textwidth}{\begin{center}
\tablestyle{5pt}{1.1}
\begin{tabular}{lcccc}
  \shline
  $\lambda ~ \downarrow ~ p ~\rightarrow$& 0 & 1 & 3 & 5 \\
  \hline
  0.0 & 26.8 & 27.4 & 27.9 & 27.7 \\
  0.01 & 26.8 & 26.8 & 27.6 & \textbf{28.3} \\
  0.1 & 26.3 & 26.8 & 27.2 & 28.0 \\
  \shline
  % \\
\end{tabular}
\end{center}}\end{minipage}
}
\hspace{1em}
\subfloat[Multi-modal embedding dimension.]{
\begin{minipage}{0.3\textwidth}{\begin{center}
\tablestyle{5pt}{1.1}
\begin{tabular}{lc}
  \shline
   Dimension & mIoU \\
  \hline
   % 256 & 27.4 \\
   384 & 28.3 \\
   512 & 28.6 \\
   1024 & \textbf{28.8} \\
  \shline
\end{tabular}
\end{center}}\end{minipage}
}
\\
\subfloat[Ground truth similarity score.]{
\begin{minipage}{0.3\textwidth}{\begin{center}
\tablestyle{5pt}{1.1}
\begin{tabular}{lc}
  \shline
   Similarity & mIoU \\
  \hline
   identity & \textbf{28.8} \\
   tf-idf & 28.4 \\
  \shline
  & \\
\end{tabular}
\end{center}}\end{minipage}
}
\subfloat[Dataset size.]{
\begin{minipage}{0.3\textwidth}{\begin{center}
\tablestyle{5pt}{1.1}
\begin{tabular}{lc}
  \shline
  Dataset Size & mIoU \\
  \hline
  10\% & 16.2 \\
  50\% & 25.3 \\
  100\% & 28.8 \\
  \shline
\end{tabular}
\end{center}}\end{minipage}
}
\hspace{1em}
\subfloat[Pretraining dataset.]{
\begin{minipage}{0.3\textwidth}{\begin{center}
\tablestyle{5pt}{1.1}
\begin{tabular}{lc}
  \shline
  Dataset & mIoU \\
  \hline
  ImageNet & 28.8 \\
  COCO & \textbf{31.7} \\
  \shline
   &  \\
\end{tabular}
\end{center}}\end{minipage}
}
\caption{Ablations of TSEG with ViT-S/16 as the image encoder and Bert-Small as the language model on PhraseCut validation set.}
\vspace{-1cm}
\label{tab:ablations}
\end{table*}

%% file: figures/methods.tex
\begin{figure*}[t]
  \centering
  % \vspace{-0.4cm}
  \includegraphics[width=\textwidth]{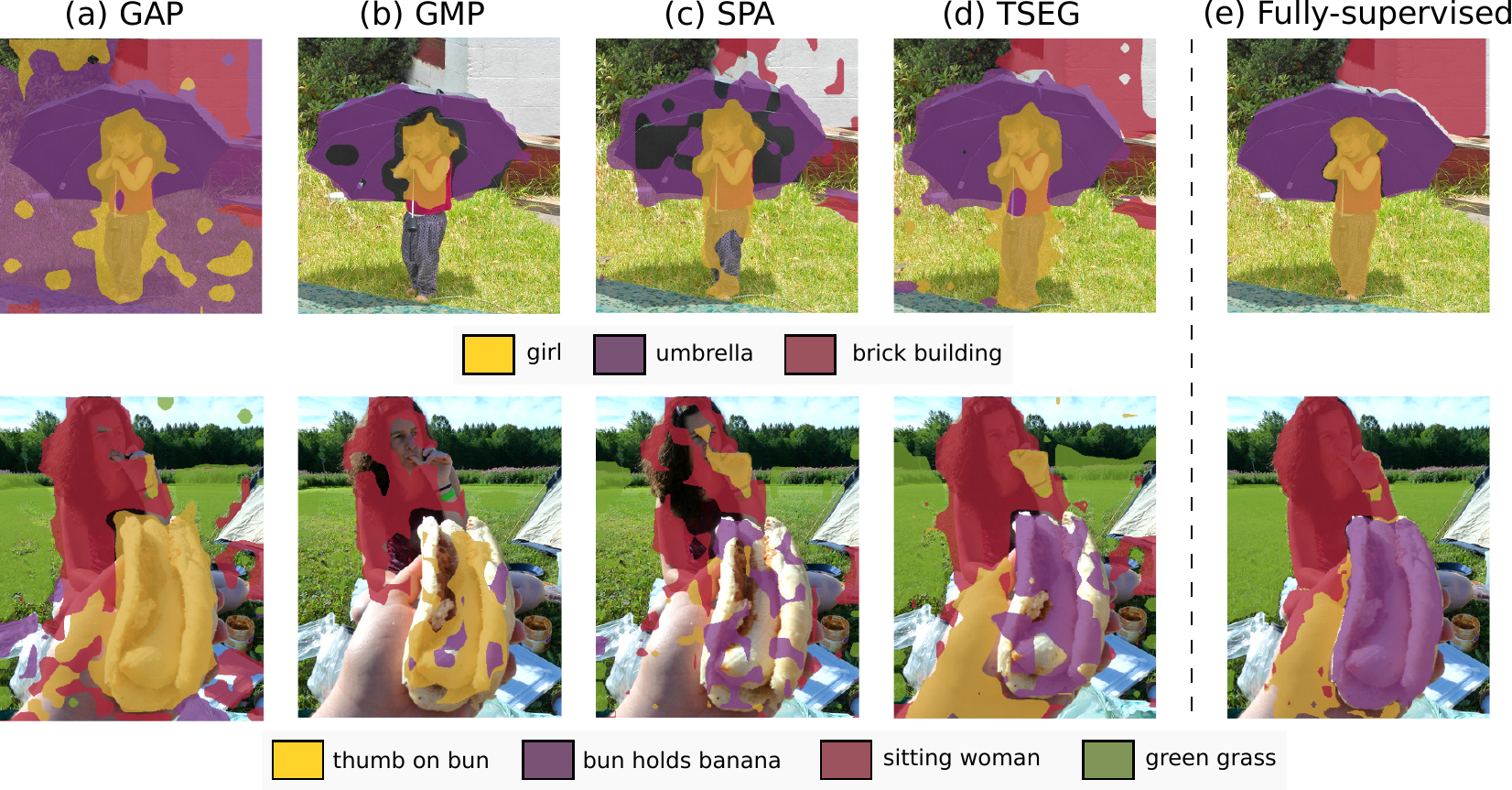}
  % \vspace{-0.4cm}
  \caption{Comparison of different pooling mechanisms for weakly supervised segmentation from referring expressions on example images from the PhraseCut dataset:
  (a) Global average pooling (GAP), (b) Global max pooling (GMP),
  (c)~Single-label patch assignment (SPA), (d) TSEG with multi-label patch assignment (TSEG). (e) Fully supervised results.}
  \label{fig:methods}
  \vspace{-0.4cm}
\end{figure*}

%% file: tables/wseg_re.tex
\begin{table*}[t]
\small
\setlength\tabcolsep{2.7pt}
\centering
% \vspace{-1.5cm}
\begin{tabular}{lcccc}
\toprule
  Method & PhraseCut & RefCOCO & RefCOCO+ & RefCOCOg \\
  \midrule
  GMP \cite{zhou16}\tsdagger & 5.77 & 6.54 & 5.12 & 6.54 \\ % & - \\
  GAP \cite{ahn18}\tsdagger & 9.35 & 6.65 & 7.21 & 6.07 \\ % & - \\
  SPA \cite{araslanov20}\tsdagger  & 21.12 & 10.32 & 9.16 & 8.35 \\
  \midrule
  TSEG & 28.77 & 25.44 & 22.01 & 22.05 \\
  TSEG (CRF) & \textbf{30.12} & \textbf{25.95} & \textbf{22.62} & \textbf{23.41} \\
\bottomrule
\end{tabular}
\caption{Comparison of different weakly-supervised methods for referring expression segmentation on
  Phrasecut and RefCOCO validation set. \tsdagger \, denotes our implementation,
  validated in Table \ref{tab:wseg_pascal}.}
\label{tab:wseg_re}
% \vspace{-0.5cm}
\end{table*}

%%% Local Variables:
%%% mode: latex
%%% TeX-master: "../main"
%%% End:

%% file: figures/qualitative.tex
\begin{figure*}[t]
  \centering
  \vspace{-0.2cm}
  \includegraphics[width=\textwidth]{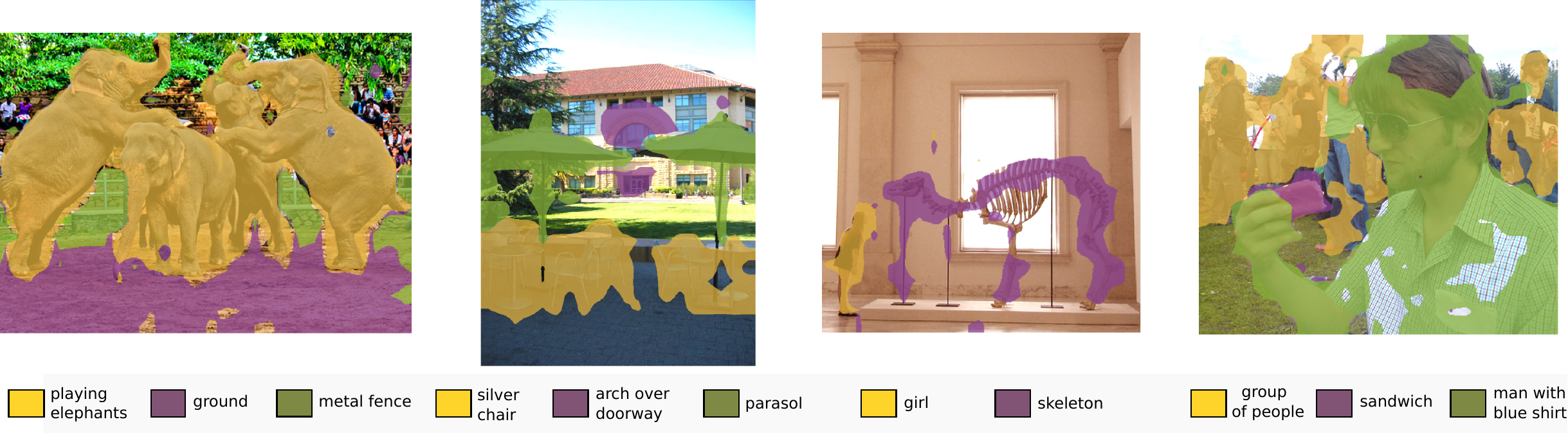}
  \caption{TSEG segmentation results on the PhraseCut test set. Our method segments
    a rich set of open-vocabulary concepts without using pixel-level supervision at the training.}
  \label{fig:qualitative}
  \vspace{-0.5cm}
\end{figure*}

% image indices
% 334
% 2323

%% file: figures/zero_shot_pascal.tex
\begin{figure*}[t]
\vspace{-0.4cm}
  \centering
  \includegraphics[width=\textwidth]{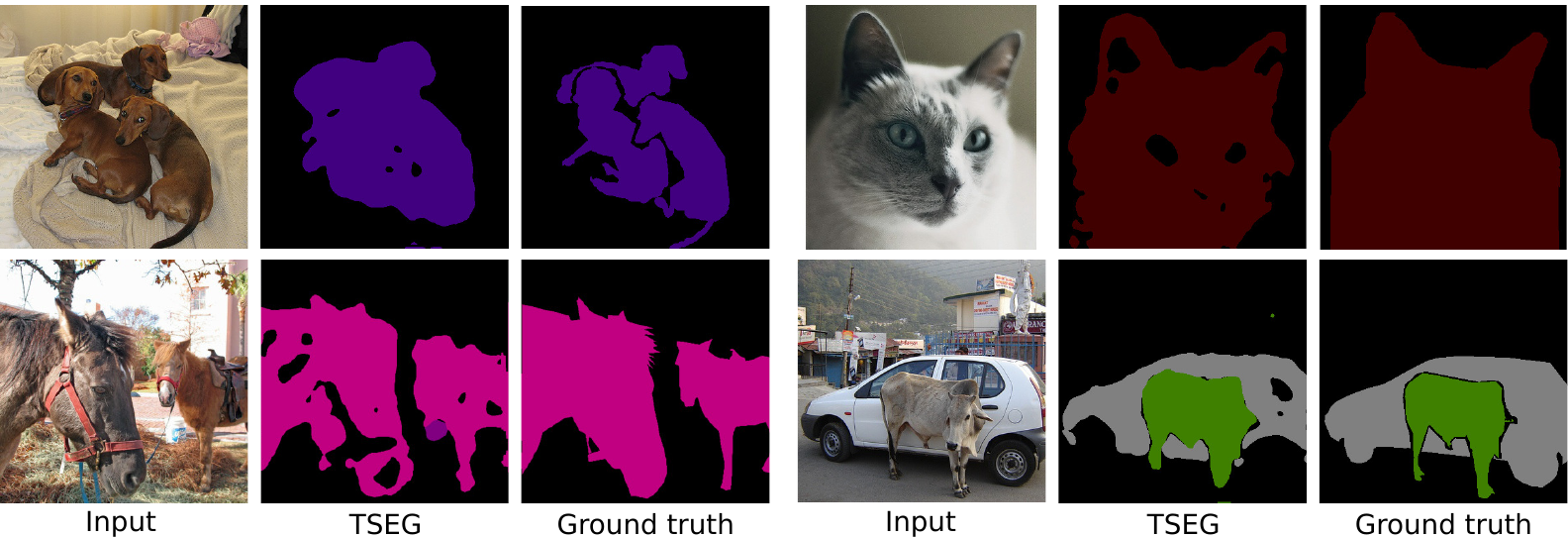}
  \caption{Zero-shot transfer of our approach TSEG trained from text supervision
    on PhraseCut and evaluated on Pascal VOC 2012. The method has not been explicitly trained for PASCAL classes and has never obtained pixel-level supervision.}
  \label{fig:zero_shot}
\end{figure*}

%% file: figures/zero_shot_fail.tex
\begin{figure*}[t]
% \vspace{-0.4cm}
  \centering
  \includegraphics[width=0.6\textwidth]{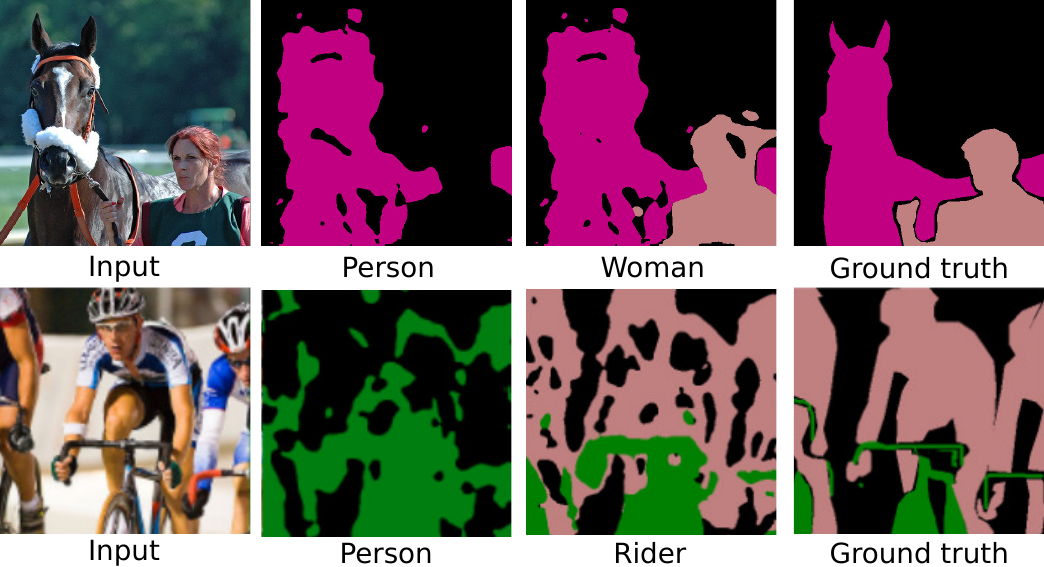}
  \caption{Failure cases on the person class for zero-shot results on Pascal
    VOC 2012. While the horse (violet) or bicycle (green) are well localized, the class person (pink) is not  detected with the \textit{person} label (column 2). The model detects it by using 
    more specific labels such as \textit{rider} or \textit{woman} (column 3, pink). Column 4 shows the ground truth.}
  \label{fig:zero_shot_fail}
\vspace{-0.4cm}
\end{figure*}

%% file: supmat/weak_vs_full.tex
\begin{figure*}[h]
  \centering
  \includegraphics[width=\textwidth]{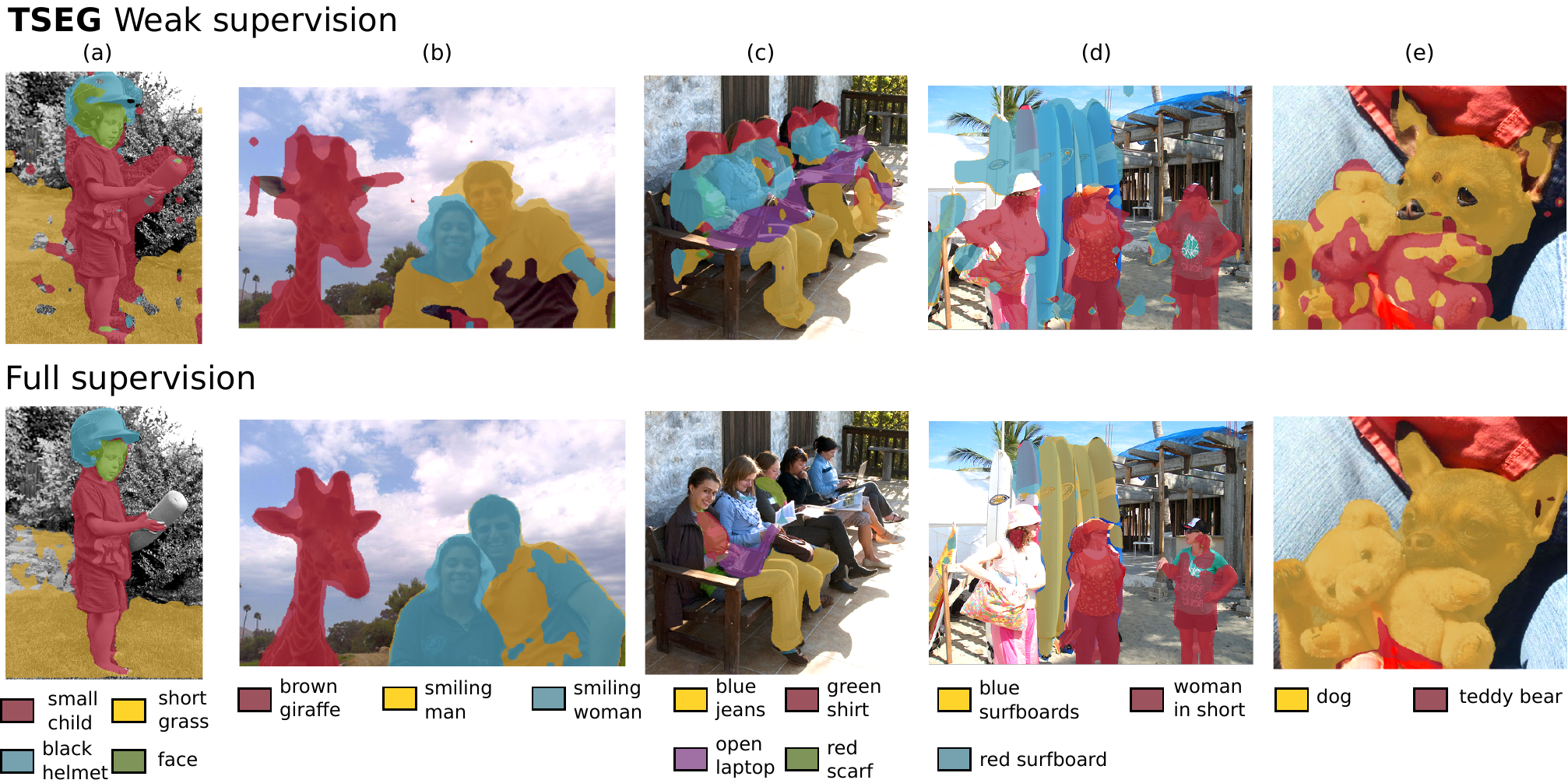}
  \caption{Comparison of TSEG to fully-supervised results on PhraseCut
  validation set. (a) Both methods perform well. (b) Both
  approaches do not distinguish well man and woman. (c-d) TSEG captures
  coarse semantic meaning such as legs (c) or surfboards (d) but misses the
  difference between a book and a laptop (c) or color attributes (d). (e) TSEG
  distinguishes the teddy bear and dog better than the fully-supervised model.}
  \label{fig:supmat_weak_vs_full}
  \vspace{-1.0cm}
\end{figure*}

%%% Local Variables:
%%% mode: latex
%%% TeX-master: "../main"
%%% End:

%% file: supmat/weak.tex
\begin{figure*}[h]
  \centering
  \includegraphics[width=\textwidth]{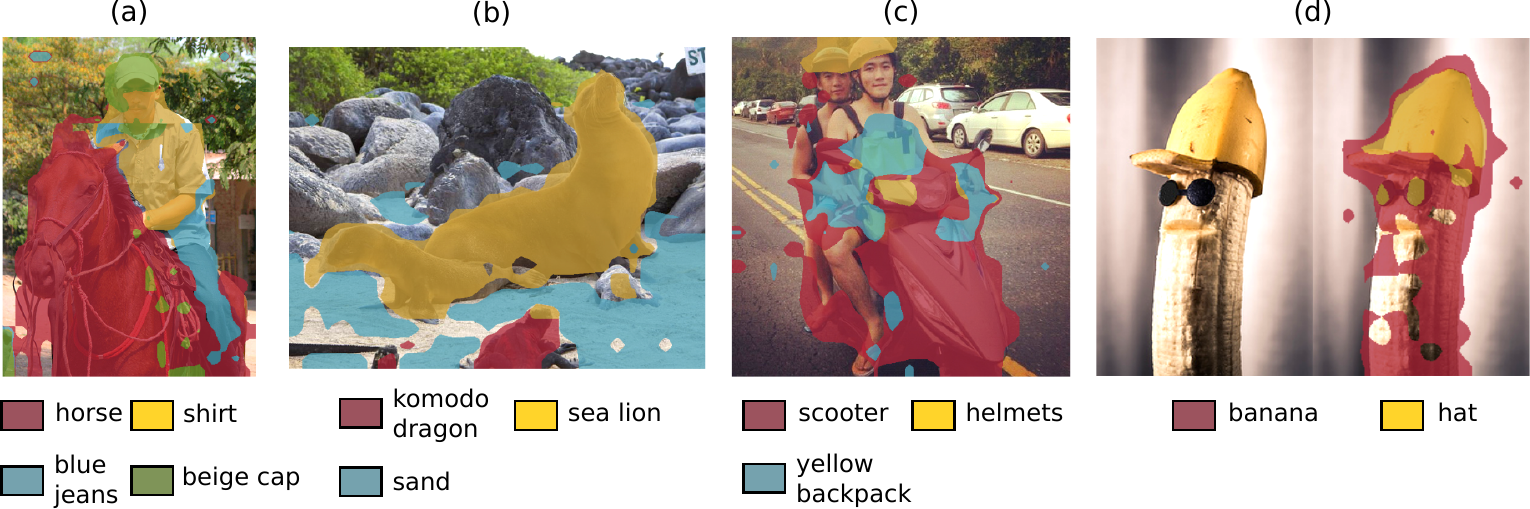}
  \caption{Additional qualitative results of our approach TSEG. Our approach
    captures rarely occurring visual concepts such as a komodo dragon or a
    banana-made hat.}
  \label{fig:supmat_weak}
  \vspace{-2.0cm}
\end{figure*}

%%% Local Variables:
%%% mode: latex
%%% TeX-master: "../main"
%%% End:

%% file: main.bbl
\begin{thebibliography}{10}
\providecommand{\url}[1]{\texttt{#1}}
\providecommand{\urlprefix}{URL }
\providecommand{\doi}[1]{https://doi.org/#1}

\bibitem{ahn19}
Ahn, J., Cho, S., Kwak, S.: Weakly supervised learning of instance segmentation
  with inter-pixel relations. In: {CVPR} (2019)

\bibitem{ahn18}
Ahn, J., Kwak, S.: Learning pixel-level semantic affinity with image-level
  supervision for weakly supervised semantic segmentation. In: {CVPR} (2018)

\bibitem{araslanov20}
Araslanov, N., Roth, S.: Single-stage semantic segmentation from image labels.
  In: {CVPR} (2020)

\bibitem{arnab21}
Arnab, A., Dehghani, M., Heigold, G., Sun, C., Lucic, M., Schmid, C.:
  Vi{V}i{T}: {A} video vision transformer. {ICCV}  (2021)

\bibitem{bilen2014weakly}
Bilen, H., Pedersoli, M., Tuytelaars, T.: Weakly supervised object detection
  with posterior regularization. In: {BMVC} (2014)

\bibitem{bojanowski2014weakly}
Bojanowski, P., Lajugie, R., Bach, F., Laptev, I., Ponce, J., Schmid, C.,
  Sivic, J.: Weakly supervised action labeling in videos under ordering
  constraints. In: {ECCV} (2014)

\bibitem{chen18_knowledge}
Chen, K., Gao, J., Nevatia, R.: Knowledge aided consistency for weakly
  supervised phrase grounding. In: {CVPR} (2018)

\bibitem{chen_deeplab}
Chen, L., Papandreou, G., Kokkinos, I., Murphy, K., Yuille, A.L.: Deeplab:
  Semantic image segmentation with deep convolutional nets, atrous convolution,
  and fully connected crfs. {IEEE} Trans. Pattern Anal. Mach. Intell.
  \textbf{40}(4),  834--848 (2018)

\bibitem{chen18}
Chen, L., Zhu, Y., Papandreou, G., Schroff, F., Adam, H.: Encoder-decoder with
  atrous separable convolution for semantic image segmentation. In: {ECCV}
  (2018)

\bibitem{chen20}
Chen, T., Kornblith, S., Norouzi, M., Hinton, G.E.: A simple framework for
  contrastive learning of visual representations. In: {ICML} (2020)

\bibitem{cheng21}
Cheng, B., Schwing, A.G., Kirillov, A.: Per-pixel classification is not all you
  need for semantic segmentation. In: {NIPS} (2021)

\bibitem{devlin19}
Devlin, J., Chang, M., Lee, K., Toutanova, K.: {BERT:} pre-training of deep
  bidirectional transformers for language understanding. In: NAACL-HLT (2019)

\bibitem{ding21}
Ding, H., Liu, C., Wang, S., Jiang, X.: Vision-language transformer and query
  generation for referring segmentation. In: {ICCV} (2021)

\bibitem{doersch2015unsupervised}
Doersch, C., Gupta, A., Efros, A.A.: Unsupervised visual representation
  learning by context prediction. In: {ICCV} (2015)

\bibitem{vit20}
Dosovitskiy, A., Beyer, L., Kolesnikov, A., Weissenborn, D., Zhai, X.,
  Unterthiner, T., Dehghani, M., Minderer, M., Heigold, G., Gelly, S.,
  Uszkoreit, J., Houlsby, N.: An image is worth 16x16 words: Transformers for
  image recognition at scale. In: {ICLR} (2021)

\bibitem{everingham10}
Everingham, M., Gool, L.V., Williams, C.K.I., Winn, J.M., Zisserman, A.: The
  pascal visual object classes {(VOC)} challenge. {IJCV}  \textbf{88}(2),
  303--338 (2010)

\bibitem{fan19}
Fan, R., Cheng, M., Hou, Q., Mu, T., Wang, J., Hu, S.: S4{N}et: Single stage
  salient-instance segmentation. In: {CVPR} (2019)

\bibitem{fan18}
Fan, R., Hou, Q., Cheng, M., Yu, G., Martin, R.R., Hu, S.: Associating
  inter-image salient instances for weakly supervised semantic segmentation.
  In: {ECCV} (2018)

\bibitem{ghadiyaram2019large}
Ghadiyaram, D., Tran, D., Mahajan, D.: Large-scale weakly-supervised
  pre-training for video action recognition. In: {CVPR} (2019)

\bibitem{ghiasi21}
Ghiasi, G., Gu, X., Cui, Y., Lin, T.: Open-vocabulary image segmentation. CoRR
  (2021)

\bibitem{gupta20}
Gupta, T., Vahdat, A., Chechik, G., Yang, X., Kautz, J., Hoiem, D.: Contrastive
  learning for weakly supervised phrase grounding. In: {ECCV} (2020)

\bibitem{hariharan11}
Hariharan, B., Arbelaez, P., Bourdev, L.D., Maji, S., Malik, J.: Semantic
  contours from inverse detectors. In: {ICCV} (2011)

\bibitem{he17}
He, K., Gkioxari, G., Doll{\'{a}}r, P., Girshick, R.B.: Mask {R-CNN}. In:
  {ICCV} (2017)

\bibitem{hu16}
Hu, R., Rohrbach, M., Darrell, T.: Segmentation from natural language
  expressions. In: {ECCV} (2016)

\bibitem{hu20}
Hu, Z., Feng, G., Sun, J., Zhang, L., Lu, H.: Bi-directional relationship
  inferring network for referring image segmentation. In: {CVPR} (2020)

\bibitem{huang16}
Huang, G., Sun, Y., Liu, Z., Sedra, D., Weinberger, K.Q.: Deep networks with
  stochastic depth. In: {ECCV} (2016)

\bibitem{huang18_dsrg}
Huang, Z., Wang, X., Wang, J., Liu, W., Wang, J.: Weakly-supervised semantic
  segmentation network with deep seeded region growing. In: {CVPR} (2018)

\bibitem{jia21}
Jia, C., Yang, Y., Xia, Y., Chen, Y., Parekh, Z., Pham, H., Le, Q.V., Sung, Y.,
  Li, Z., Duerig, T.: Scaling up visual and vision-language representation
  learning with noisy text supervision. In: {ICML} (2021)

\bibitem{kamath21}
Kamath, A., Singh, M., LeCun, Y., Misra, I., Synnaeve, G., Carion, N.: {MDETR}
  - modulated detection for end-to-end multi-modal understanding. {ICCV}
  (2021)

\bibitem{kantorov2016contextlocnet}
Kantorov, V., Oquab, M., Cho, M., Laptev, I.: Contextlocnet: Context-aware deep
  network models for weakly supervised localization. In: {ECCV} (2016)

\bibitem{kim21}
Kim, B., Han, S., Kim, J.: Discriminative region suppression for
  weakly-supervised semantic segmentation. In: {AAAI} (2021)

\bibitem{kingma14}
Kingma, D.P., Ba, J.: Adam: {A} method for stochastic optimization. In: {ICLR}
  (2015)

\bibitem{kolesnikov16}
Kolesnikov, A., Lampert, C.H.: Seed, expand and constrain: Three principles for
  weakly-supervised image segmentation. In: {ECCV} (2016)

\bibitem{krishna17}
Krishna, R., Zhu, Y., Groth, O., Johnson, J., Hata, K., Kravitz, J., Chen, S.,
  Kalantidis, Y., Li, L., Shamma, D.A., Bernstein, M.S., Fei{-}Fei, L.: Visual
  genome: Connecting language and vision using crowdsourced dense image
  annotations. {IJCV}  \textbf{123}(1),  32--73 (2017)

\bibitem{lee19_iccv}
Lee, J., Kim, E., Lee, S., Lee, J., Yoon, S.: Frame-to-frame aggregation of
  active regions in web videos for weakly supervised semantic segmentation. In:
  {ICCV} (2019)

\bibitem{lee21}
Lee, S., Lee, M., Lee, J., Shim, H.: Railroad is not a train: Saliency as
  pseudo-pixel supervision for weakly supervised semantic segmentation. In:
  {CVPR} (2021)

\bibitem{li2016weakly}
Li, D., Huang, J.B., Li, Y., Wang, S., Yang, M.H.: Weakly supervised object
  localization with progressive domain adaptation. In: {CVPR} (2016)

\bibitem{li19}
Li, L.H., Yatskar, M., Yin, D., Hsieh, C., Chang, K.: Visual{BERT}: {A} simple
  and performant baseline for vision and language. arXiv preprint
  arXiv:1908.03557  (2019)

\bibitem{lin20}
Lin, T., Goyal, P., Girshick, R.B., He, K., Doll{\'{a}}r, P.: Focal loss for
  dense object detection. {IEEE} Trans. Pattern Anal. Mach. Intell.
  \textbf{42}(2),  318--327 (2020)

\bibitem{liu19}
Liu, X., Li, L., Wang, S., Zha, Z., Meng, D., Huang, Q.: Adaptive
  reconstruction network for weakly supervised referring expression grounding.
  In: {ICCV} (2019)

\bibitem{liu21}
Liu, Y., Wan, B., Ma, L., He, X.: Relation-aware instance refinement for weakly
  supervised visual grounding. In: {CVPR} (2021)

\bibitem{liu21_swin}
Liu, Z., Lin, Y., Cao, Y., Hu, H., Wei, Y., Zhang, Z., Lin, S., Guo, B.: Swin
  transformer: Hierarchical vision transformer using shifted windows. In:
  {ICCV} (2021)

\bibitem{loshchilov19}
Loshchilov, I., Hutter, F.: Decoupled weight decay regularization. In: {ICLR}
  (2019)

\bibitem{miech2020end}
Miech, A., Alayrac, J.B., Smaira, L., Laptev, I., Sivic, J., Zisserman, A.:
  End-to-end learning of visual representations from uncurated instructional
  videos. In: {CVPR} (2020)

\bibitem{papandreou15}
Papandreou, G., Chen, L., Murphy, K.P., Yuille, A.L.: Weakly-and
  semi-supervised learning of a deep convolutional network for semantic image
  segmentation. In: {ICCV} (2015)

\bibitem{pinheiro15}
Pinheiro, P.H.O., Collobert, R.: From image-level to pixel-level labeling with
  convolutional networks. In: {CVPR} (2015)

\bibitem{radford21}
Radford, A., Kim, J.W., Hallacy, C., Ramesh, A., Goh, G., Agarwal, S., Sastry,
  G., Askell, A., Mishkin, P., Clark, J., Krueger, G., Sutskever, I.: Learning
  transferable visual models from natural language supervision. In: {ICML}
  (2021)

\bibitem{ramesh21}
Ramesh, A., Pavlov, M., Goh, G., Gray, S., Voss, C., Radford, A., Chen, M.,
  Sutskever, I.: Zero-shot text-to-image generation. In: {ICML} (2021)

\bibitem{ren17}
Ren, S., He, K., Girshick, R.B., Sun, J.: Faster {R-CNN:} towards real-time
  object detection with region proposal networks. {PAMI}  \textbf{39}(6),
  1137--1149 (2017)

\bibitem{robbins1951}
Robbins, H., Monro, S.: A stochastic approximation method. Annals of
  Mathematical Statistics  (1951)

\bibitem{bpe16}
Sennrich, R., Haddow, B., Birch, A.: Neural machine translation of rare words
  with subword units. In: {ACL} (2016)

\bibitem{steiner21}
Steiner, A., Kolesnikov, A., Zhai, X., Wightman, R., Uszkoreit, J., Beyer, L.:
  How to train your {V}i{T}? data, augmentation, and regularization in vision
  transformers. arXiv preprint arXiv:2106.10270  (2021)

\bibitem{strudel21}
Strudel, R., Pinel, R.G., Laptev, I., Schmid, C.: Segmenter: Transformer for
  semantic segmentation. {ICCV}  (2021)

\bibitem{turc19}
Turc, I., Chang, M., Lee, K., Toutanova, K.: Well-read students learn better:
  The impact of student initialization on knowledge distillation. arXiv
  preprint arXiv:1908.08962  (2019)

\bibitem{vaswani17}
Vaswani, A., Shazeer, N., Parmar, N., Uszkoreit, J., Jones, L., Gomez, A.N.,
  Kaiser, L., Polosukhin, I.: Attention is all you need. In: {NIPS} (2017)

\bibitem{wang17}
Wang, J., Jiang, H., Yuan, Z., Cheng, M., Hu, X., Zheng, N.: Salient object
  detection: {A} discriminative regional feature integration approach. {IJCV}
  \textbf{123}(2),  251--268 (2017)

\bibitem{wei17}
Wei, Y., Feng, J., Liang, X., Cheng, M., Zhao, Y., Yan, S.: Object region
  mining with adversarial erasing: {A} simple classification to semantic
  segmentation approach. In: {CVPR} (2017)

\bibitem{wei18}
Wei, Y., Xiao, H., Shi, H., Jie, Z., Feng, J., Huang, T.S.: Revisiting dilated
  convolution: {A} simple approach for weakly- and semi-supervised semantic
  segmentation. In: {CVPR} (2018)

\bibitem{wu20}
Wu, C., Lin, Z., Cohen, S., Bui, T., Maji, S.: Phrasecut: Language-based image
  segmentation in the wild. In: {CVPR} (2020)

\bibitem{xiao17}
Xiao, F., Sigal, L., Lee, Y.J.: Weakly-supervised visual grounding of phrases
  with linguistic structures. In: {CVPR} (2017)

\bibitem{xu22}
Xu, J., Mello, S.D., Liu, S., Byeon, W., Breuel, T.M., Kautz, J., Wang, X.:
  Groupvit: Semantic segmentation emerges from text supervision. CoRR  (2022)

\bibitem{xu21}
Xu, M., Zhang, Z., Wei, F., Lin, Y., Cao, Y., Hu, H., Bai, X.: A simple
  baseline for zero-shot semantic segmentation with pre-trained vision-language
  model. CoRR  (2021)

\bibitem{ye19}
Ye, L., Rochan, M., Liu, Z., Wang, Y.: Cross-modal self-attention network for
  referring image segmentation. In: {CVPR} (2019)

\bibitem{yu18}
Yu, L., Lin, Z., Shen, X., Yang, J., Lu, X., Bansal, M., Berg, T.L.: Mattnet:
  Modular attention network for referring expression comprehension. In: {CVPR}
  (2018)

\bibitem{yu16}
Yu, L., Poirson, P., Yang, S., Berg, A.C., Berg, T.L.: Modeling context in
  referring expressions. In: {ECCV} (2016)

\bibitem{yu19}
Yu, Z., Zhuge, Y., Lu, H., Zhang, L.: Joint learning of saliency detection and
  weakly supervised semantic segmentation. In: {ICCV} (2019)

\bibitem{zabari21}
Zabari, N., Hoshen, Y.: Semantic segmentation in-the-wild without seeing any
  segmentation examples. CoRR  (2021)

\bibitem{zhou16}
Zhou, B., Khosla, A., Lapedriza, {\`{A}}., Oliva, A., Torralba, A.: Learning
  deep features for discriminative localization. In: {CVPR} (2016)

\bibitem{zhou21_denseclip}
Zhou, C., Loy, C.C., Dai, B.: Denseclip: Extract free dense labels from {CLIP}.
  CoRR  (2021)

\end{thebibliography}
